\documentclass[10pt,twocolumn,letterpaper]{article}

\usepackage{cvpr}
\usepackage{times}
\usepackage{epsfig}
\usepackage{graphicx}
\usepackage{amsmath}
\usepackage{amssymb}

\usepackage{subcaption}

\usepackage[pagebackref=true,breaklinks=true,letterpaper=true,colorlinks,bookmarks=false]{hyperref}

\cvprfinalcopy % *** Uncomment this line for the final submission

\ifcvprfinal\fi
\begin{document}

%%%%%%%%% TITLE
\title{Learning Image Conditioned Label Space for Multilabel Classification}

\author{Yi-Nan Li and Mei-Chen Yeh\\
Department of Computer Science and Information Engineering\\
National Taiwan Normal University\\
{\tt\small myeh@csie.ntnu.edu.com }
% For a paper whose authors are all at the same institution,
% omit the following lines up until the closing ``}''.
% Additional authors and addresses can be added with ``\and'',
% just like the second author.
% To save space, use either the email address or home page, not both
%\and
%Mei-Chen Yeh\\
%Department of Computer Science and Information Engineering\\
%National Taiwan Normal University\\
%{\tt\small myeh@csie.ntnu.edu.tw}
}

\maketitle
%\thispagestyle{empty}

%%%%%%%%% ABSTRACT
\begin{abstract}

This work addresses the task of multilabel image classification. Inspired by the great success from deep convolutional neural networks (CNNs) for single-label visual-semantic embedding, we exploit extending these models for multilabel images. Specifically, we propose an image-dependent ranking model, which returns a ranked list of labels according to its relevance to the input image. In contrast to conventional CNN models that learn an image representation (i.e. the image embedding vector), the developed model learns a mapping (i.e. a transformation matrix) from an image in an attempt to differentiate between its relevant and irrelevant labels. Despite the conceptual simplicity of our approach, experimental results on a public benchmark dataset demonstrate that the proposed model achieves state-of-the-art performance while using fewer training images than other multilabel classification methods.

\end{abstract}

%%%%%%%%% BODY TEXT

\section{Introduction}

\label{sec:intro}

\begin{figure}[t]
\begin{center}
%\fbox{\rule{0pt}{2in} \rule{0.9\linewidth}{0pt}}
 \includegraphics[width=0.95\linewidth]{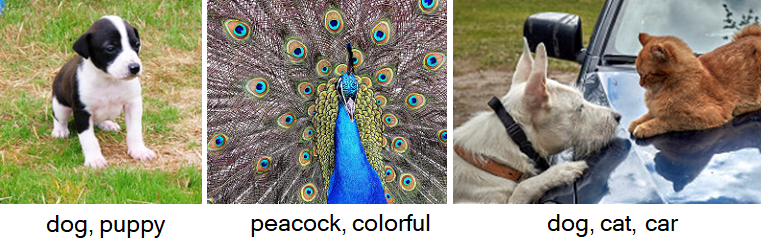}
\end{center}
   \caption{An image is often annotated with several tags: (left) semantically similar, (middle) and (right) semantically different.}
\label{fig:intro}
\end{figure}

%Multi-label image classification ~\cite{Tagprop, Baselines, Wsabie, Pascal} is an important problem in computer vision, in which multiple labels are assigned to one image based on its content. Compared to single-label image classification, the task is more general, yet, also a more challenging one because of the rich semantic information and complex dependency of an image and its labels. For example, image labels may have overlapping meanings. \textit{Car} and \textit{vehicle} have similar meanings and are often interchangeable (Figure \ref{fig:intro} (left)). Meanwhile, labels may be semantically different, capturing one (Figure \ref{fig:intro} (middle)) or multiple (Figure \ref{fig:intro} (right)) objects in the image. These labels may exhibit strong co-occurrence dependencies; for example, \textit{sky} and \textit{clouds} are semantically different, but often appear together on one image. 
Multilabel image classification~\cite{Tagprop, Baselines, Wsabie, Pascal} is a crucial problem in computer vision, where the goal is to assign multiple labels to one image based on its content. Compared with single-label image classification, multilabel image classification is more general, but it is also more challenging because of the rich semantic information and complex dependency of an image and its labels. For example, image labels may have overlapping meanings. \textit{Dog} and \textit{puppy} have similar meanings and are often interchangeable (Figure \ref{fig:intro} (left)). Moreover, labels may be semantically different, capturing one (Figure \ref{fig:intro} (middle)) or multiple (Figure \ref{fig:intro} (right)) objects in the image. Such labels may exhibit strong co-occurrence dependencies; for example, \textit{sky} and \textit{clouds} are semantically different, but they often appear together in one image.

%The current state-of-the-art approach to image classification is a deep convolutional neural network (CNN) trained with a softmax output layer (i.e. multinomial logistic regression) that has as many units as the number of classes ~\cite{AlexNet}. A common approach that extends CNN to multi-label classification is to transform it into multiple single-label classification problems, which can be trained with the ranking loss~\cite{WARP_0} or the cross-entropy loss~\cite{Tagprop}. However, as the number of classes grows, the distinction between classes blurs, and it becomes increasingly difficult to obtain sufficient numbers of training images for rare concepts. While the number of labels continuing to grow, these models are often limited in their ability to scale to large numbers of object categories (introducing many model parameters, difficult to obtain sufficient numbers of training images). Furthermore, these methods fail to model the dependency between labels.
The current state-of-the-art approach to image classification is a deep convolutional neural network (CNN) trained with a softmax output layer (i.e. multinomial logistic regression) that has as many units as the number of classes~\cite{AlexNet}. A common approach to extending CNN to multilabel classification is to transform it into multiple single-label classification problems, which can be trained with the ranking loss~\cite{WARP_0} or the cross-entropy loss~\cite{Tagprop}. However, as the number of classes grows, the distinction between classes is obscured, and it becomes increasingly difficult to obtain sufficient numbers of training images for rare concepts. As the number of labels continues to grow, these models are often limited in their scalability to large numbers of object categories (introducing many model parameters, making it difficult to obtain sufficient numbers of training images). Furthermore, these methods fail to model the dependency between labels.

\begin{figure}[t]
\begin{center}
%\fbox{\rule{0pt}{2in} \rule{0.9\linewidth}{0pt}}
 \includegraphics[width=0.99\linewidth]{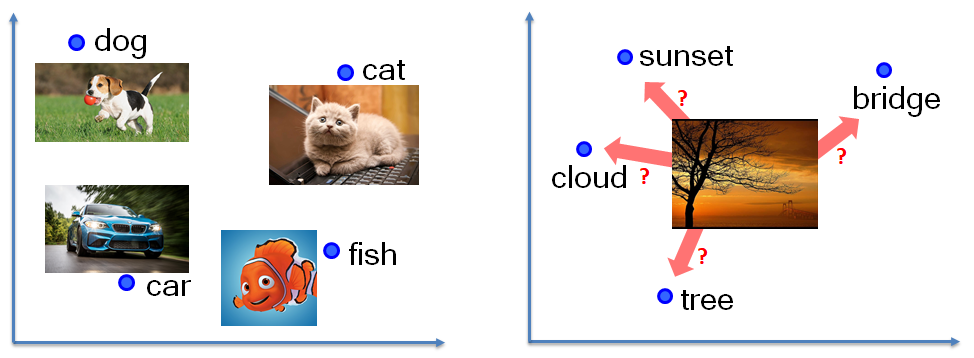}
\end{center}
	  \caption{Visual-semantic embedding maps images into a semantic label space. This task is trivial for single-label images (left), but not the case for multilabel images (right).}
\label{fig:embedding}
\end{figure}

%Alternatively, visual-semantic embedding models address these shortcomings by training a visual recognition model with both labeled images and a large corpus of unannotated text~\cite{DeViSE_0, zero-shot}. Textual data are leveraged to learn semantic relationships between labels, in which semantically similar labels are close to each other in the continuous embedding space. An image is transformed into that space and is close to its associated labels. While these image embedding methods have stressed their advantages over traditional $n$-way classifiers, handing images with multiple labels still remains an open problem, because an image may contain more than one semantic concept, as depicted in Figure \ref{fig:embedding}.
Alternatively, visual-semantic embedding models address these shortcomings by training a visual recognition model with both labeled images and a large corpus of unannotated text~\cite{DeViSE_0, zero-shot}. Textual data are leveraged to learn semantic relationships between labels, with semantically similar labels being close to each other in the continuous embedding space. An image is transformed into that space and is close to its associated labels. Although the advantages of these image embedding methods over traditional $n$-way classifiers have been highlighted, handling images with multiple labels still remains problematic, because an image may contain more than one semantic concept, as depicted in Figure \ref{fig:embedding}.

The characteristic of varying and unordered labels one image may have in multilabel image classification hinders the direct employment of CNN that requires a fixed output size. Wei \textit{et al.}~\cite{HCP} tackled the problem by creating an arbitrary number of object segment hypotheses as the inputs to a shared CNN. However, the classification performance depends largely on the quality of the extracted hypotheses and an ideal way to extract them remains unclear.  Wang \textit{et al.}~\cite{CNN-RNN_0} proposed CNN-RNN, which utilized recurrent neural networks (RNNs)~\cite{LSTM} to address this problem. Although the recurrent neurons neatly model the label co-occurrence dependencies, this approach needs to determine the orders of the labels from an unordered label set. Both methods significantly increase the model complexity (e.g. computing hypotheses or integrating RNNs) to extend CNN from single-label to multilabel image classification.

\begin{figure}[t]
\begin{center}
%\fbox{\rule{0pt}{2in} \rule{0.9\linewidth}{0pt}}
 \includegraphics[width=0.9\linewidth]{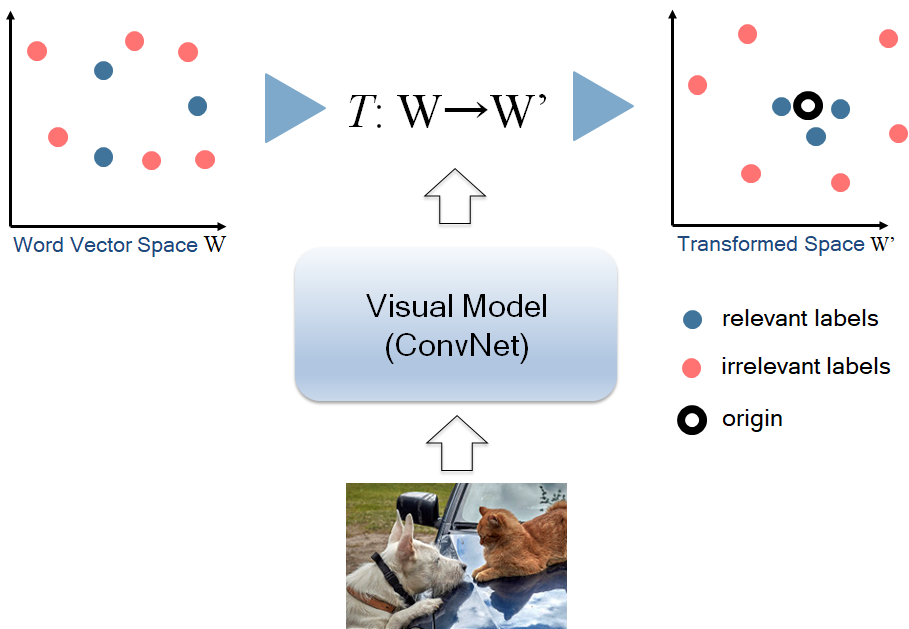}
\end{center}
   \caption{Infrastructure of the proposed CNN model. The model learns image-dependent transformation, which can be used to return multilabel predictions for a new image.}
\label{fig:framework}
\end{figure}

%In this paper, we explore to extend visual-semantic embedding models for multi-label image classification. One key observation is that, despite the complex relationship among labels in the semantic space, one image is considered as a conduit for constructing the relationship of its labels. More specifically, an image divides all words into two sets according to the image-label relevance~\cite{Fast0Tag}. Therefore, we develop an image-dependent ranking model, which returns a ranked list of labels according to its relevance to the input image. The idea is implemented using a simple CNN framework, as shown in Figure \ref{fig:framework}. Unlike conventional CNN models that learn an image representation (i.e., the image embedding vector), we learn a mapping (i.e., a transformation matrix) from an image that attempt to differentiate between its relevant and irrelevant labels. During prediction, the image transformation is used to map words from the input word space into a new space, where the words can satisfactorily be ranked according to their relevance to the input image. The proposed framework has the advantage of the visual-semantic embedding that neatly address the semantic redundancy among labels, it also models the label co-occurrence without introducing additional subnets to be integrated in the CNN framework. 
In this study, we explored and extended visual-semantic embedding models for multilabel image classification. One key observation is that, despite the complex relationship among labels in the semantic space, one image is considered as a conduit for constructing the relationship of its labels. Specifically, an image divides all words into two sets according to the image-label relevance~\cite{Fast0Tag}. Therefore, we developed an image-dependent ranking model, which returns a ranked list of labels according to its relevance to the input image. The idea was implemented using a simple CNN framework, as shown in Figure \ref{fig:framework}. In contrast to conventional CNN models that learn an image representation (i.e. the image embedding vector), the developed model learns a mapping (i.e. a transformation matrix) from an image in an attempt to differentiate between its relevant and irrelevant labels. During prediction, the image transformation matrix is used to map words from the input word space into a new space, where the words can satisfactorily be ranked according to their relevance to the input image. The proposed framework has the advantage of applying visual-semantic embedding that effectively addresses the semantic redundancy among labels; it also models the label co-occurrence without introducing additional subnets that are to be integrated in the CNN framework. 

%Compared with state-of-the-art multi-label image classification methods, the proposed CNN model has the following characteristics:
Compared with state-of-the-art multilabel image classification methods, the proposed CNN model has the following characteristics:

\begin{itemize}
  %\item The model takes an image as the input and give multi-label predictions, without the computation of object segments or local image regions, neither requires ground-truth bounding box information for training.
	\item The model takes an image as the input and provides multilabel predictions without the computation of object segments or local image regions or the requirement of ground-truth bounding box information for training.
  %\item The model addresses the semantic redundancy and the co-occurrence dependency problems in multi-label classification, and can be trained efficiently in an end-to-end manner.
	\item The model addresses the semantic redundancy and the co-occurrence dependency problems in multilabel classification, and it can be trained efficiently in an end-to-end manner.
	%\item The model learns from an image a transformation, rather than a representation. The output transformation can be readily used to solve the multi-label classification problem.
	\item The model learns from an image a transformation, rather than a representation. The output transformation can be readily used to solve the multilabel classification problem.
	%\item The model is conceptually simple and compact. Yet it is more powerful than many existing deep learning based models for multi-label classification.
	\item The model is conceptually simple and compact, yet it is more powerful than many existing deep learning-based models for multilabel classification.
\end{itemize}

%We evaluate the proposed framework with experiments on the public multi-label benchmark dataset NUS-WIDE. Experimental results demonstrate that the proposed method uses less training data but achieves better performance compared to the current state-of-the-art multi-label classification methods. We further make an attempt to explain the superior performance of the model and to empirically interpret the behavior of the model in the discussion section. In the remainder of the paper, we briefly review the related work of multi-label classification in Section \ref{sec:review}. Section \ref{sec:method} presents the details of how we use a single CNN model to extend visual-semantic embedding for multi-label image classification. The experimental results and conclusions are finally provided in Section \ref{sec:exp} and Section \ref{sec:conclusion}.
We evaluated the proposed framework with experiments conducted on the public multilabel benchmark dataset NUS-WIDE~\cite{NUS-WIDE}. Experimental results demonstrated that the proposed method uses less training data but achieves superior performance, compared with current state-of-the-art multilabel classification methods. We  further explain the superior performance of the model and empirically interpret the behavior of the model in the Discussion section. The remainder of the paper is organized as follows: We briefly review work related to multilabel classification in Section \ref{sec:review}. Section \ref{sec:method} presents the details of the processes involved in using a single CNN model to extend visual-semantic embedding for multilabel image classification. The experimental results and conclusions are provided in Sections \ref{sec:exp} and \ref{sec:conclusion}, respectively.

\section{Related Works}
\label{sec:review}

In this section, we briefly review prior work on multilabel image classification. We start with a few hand-crafted features based methods and then describe deep learning based methods.

Pioneering work for addressing the classification problem focused on learning statistical models using hand-crafted features. For example, Makadia \textit{et al.}~\cite{Baselines} utilized global low-level image features and a simple combination of basic distance measures to find nearest neighbors of a given image. The keywords were then assigned using a greedy label transfer mechanism. A bag-of-words model that aggregates local features extracted densely from an image was applied in~\cite{Harzallah, Perronnin, Chen, Dong}. In particular, Weston \textit{et al.}~\cite{Wsabie} proposed a loss function to embed images with the associated labels together based on bag-of-words. Although these works have shown some successes, hand-crafted features are not always optimal for this particular task. 

In contrast to hand-crafted features, learned features with deep learning have shown great potential for various vision recognition tasks. Specifically, CNN~\cite{CNN} has demonstrated an extraordinary ability for single-label image classification~\cite{Jarrett, LeCun, Lee, AlexNet, Lin}. To extend CNN to multilabel classification, Sermanet \textit{et al.}~\cite{OverFeat} and Razavian \textit{et al.}~\cite{Razavian} used a CNN feature-SVM pipeline, in which multilabel images were directly fed into a CNN---pretrained on ImageNet~\cite{ImageNet}---to get CNN activations as the of-the-shelf features for classification. Beyond using CNN as a feature extractor, Gong \textit{et al.}~\cite{WARP_0} compared several popular multilabel losses to train the network. Using a top-$k$ ranking objective achieved state-of-the-art performance. Li \textit{et al.}~\cite{Li} improved that objective using the log-sum-exp pairwise function. Hypotheses-CNN-Pooling~\cite{HCP} employed max pooling to aggregate the predictions from multiple hypothesis region proposals. These methods treated each label independently and ignored the semantic relationships between labels.
% A few CNN-based methods handle multiple labels by treating an image as multiple images sampled from different regions. For example, 

Visual-semantic embedding models~\cite{DeViSE_0, zero-shot} effectively exploit the label semantic redundancy by leveraging the textual data. Instead of manually designing the semantic label space, Frome \textit{et al.}~\cite{DeViSE_0} and Norouzi \textit{et al.}~\cite{zero-shot} used semantic information gleaned from unannotated text to learn visual-semantic embedding where semantic relationship between labels was preserved. To extend visual-semantic embedding models to multilabel classification, Wang \textit{et al.}~\cite{CNN-RNN_0} utilized RNNs to exploit the label dependencies in an image. The recurrent neurons model the label co-occurrence dependencies by sequentially linking the label embeddings in the joint embedding space. Similar to~\cite{HCP} in which an image is represented by a number of regions of interests, the multi-instance visual-semantic embedding model (MIE)~\cite{MIE_0} mapped the image subregions to their corresponding labels. These methods introduced significant complexity into the CNN architecture and may not be suitable for tasks that do not have powerful computing resources. 

The proposed method uses the identical infrastructure to DeViSE~\cite{DeViSE_0}, involving only a single CNN to operate visual information. Our key motivation is to design a simple method with a new modeling-paradigm, which extends DeViSE to processing multilabel images. The proposed method is fast in training and offers instant prediction during testing (only a linear transformation is required).

\section{Method}
\label{sec:method}

The objective of this study is to extend visual-semantic embedding models for multilabel image classification. We use a CNN and a word embedding model to achieve the goal. We start with formulating a binary classification problem for multilabel image classification. Next, we describe in details the model architecture and training.

\subsection{Formulation}
\label{subsec:formulation}

\begin{figure}[t]
\begin{center}
%\fbox{\rule{0pt}{2in} \rule{0.9\linewidth}{0pt}}
 \includegraphics[width=0.9\linewidth]{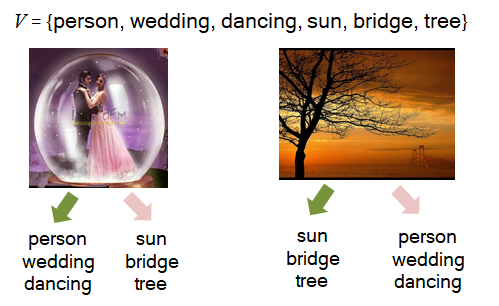}
\end{center}
   \caption{Simplified example of formulating the multilabel image classification as a binary classification problem.}
\label{fig:formulation}
\end{figure}

In this study, we consider the task of multiclass image classification as a single \textit{binary} classification problem. Figure \ref{fig:formulation} illustrates a simplified example. Suppose we have a label set $V$ containing six words: \textit{person}, \textit{wedding}, \textit{dancing}, \textit{sun}, \textit{bridge} and \textit{tree}. Figure \ref{fig:formulation} (left) separates the words into two classes, where \textit{person}, \textit{wedding} and \textit{dancing} are considered positive because these labels suitably describe the image. Similarly, Figure \ref{fig:formulation} (right) shows a different partition of words. 

Given an image, we aim at partitioning labels into two disjoint sets according to the image-label relevance. The partition $(X, V \backslash X)$ involves analyzing the relationship between an image and two sets of words. Based on this observation, we propose to learn an \textit{image-dependent} classifier, which is able to separate the relevant and irrelevant labels of an input image.

\subsection{Architecture}
\label{subsec:architecture}

\begin{figure*}[t]
\begin{center}
%\fbox{\rule{0pt}{2in} \rule{0.9\linewidth}{0pt}}
 \includegraphics[width=0.8\linewidth]{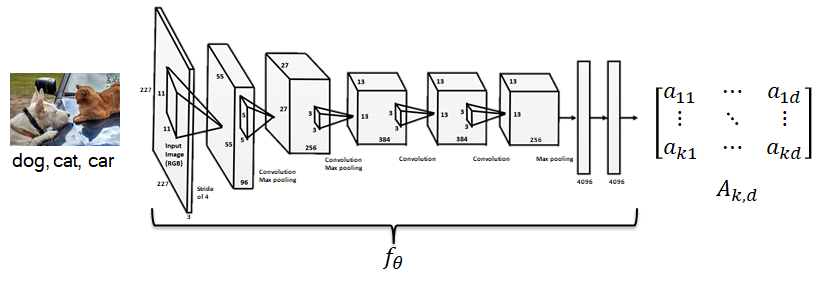}
\end{center}
   \caption{Infrastructure of the proposed CNN model. The model has a similar network structure to~\cite{AlexNet} except for the layer fc8, where the node number is equal to the size of the transformation matrix.
}
\label{fig:architecture}
\end{figure*}

The image-dependent classifier has a form of a linear transformation matrix, which is implemented using a CNN framework (shown in Figure \ref{fig:architecture}). The model architecture is similar to the network described in~\cite{AlexNet} except for the layer fc8, where the dimension is set to the size of the transformation matrix. Namely, the output of the last layer is a vector of length $k \times d$, which can also be viewed as a $k \times d$ matrix. The image-dependent transformation matrix is used to map labels from the $d$-dim word vector space into a $k$-dim Euclidean space, where the relevant and irrelevant labels can satisfactorily be separated. Table \ref{tab:spec} specifies the number of parameters used in each layer. 

In contrast to previous works~\cite{Fast_1, DeViSE_0, Wsabie, Fast0Tag}, the linear transformation is \textit{not} used to map an image to the word vector space. Instead, the transformation learned from an image seeks for linear combinations of variables in the word vectors that characterize two sets of labels.

\begin{table*}
\begin{center}
\begin{tabular}{|l|c|c|c|c|c|}
\hline
Layer	& Input	 & Kernel	& Stride	& Output	& No. of parameters \\
\hline\hline
conv1	& 3@227$\times$227	& $11\times11$	& 4	& 96@55$\times$55	  & 96$\times$(11$\times$11$\times$3+1) \\
pool1	& 96@55$\times$55	  & $3\times3$	  & 2	& 96@27$\times$27	  & 0																		\\
conv2	& 96@27$\times$27	  & $5\times5$	  & 1	& 256@27$\times$27	& 256$\times$(5$\times$5$\times$96+1) \\
pool2	& 256@27$\times$27	& $3\times3$	  & 2	& 256@13$\times$13	& 0																		\\
conv3	& 256@13$\times$13	& $3\times3$	  & 1	& 384@13$\times$13	& 384$\times$(3$\times$3$\times$256+1)\\
conv4	& 384@13$\times$13	& $3\times3$	  & 1	& 384@13$\times$13	& 384$\times$(3$\times$3$\times$384+1)\\
conv5	& 384@13$\times$13	& $3\times3$	  & 1	& 256@13$\times$13	& 256$\times$(3$\times$3$\times$384+1)\\
pool5	& 256@13$\times$13	& $3\times3$	  & 2	& 256@6$\times$6	  & 0																		\\
fc6	  & 9216@1$\times$1	  & $1\times1$	  & 1	& 4096@1$\times$1  	& 4096$\times$(9216+1)								\\
fc7	  & 4096@1$\times$1	  & $1\times1$	  & 1	& 4096@1$\times$1	  & 4096$\times$(4096+1)								\\
fc8   & 4096@1$\times$1	  & $1\times1$	  & 1	& ($k \times d$)@1$\times$1	& ($k \times d$)$\times$(4096+1) \\
\hline
\end{tabular}
\end{center}
\caption{Network parameters}
\label{tab:spec}
\end{table*}

\subsection{Loss function} 
\label{subsec:loss} 

\begin{figure}[t]
\begin{center}
%\fbox{\rule{0pt}{2in} \rule{0.9\linewidth}{0pt}}
 \includegraphics[width=0.7\linewidth]{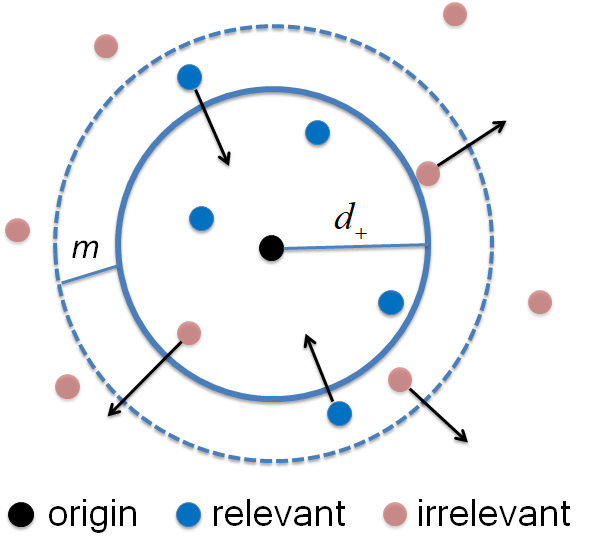}
\end{center}
   \caption{Illustration of hinge rank loss. The model attempts to map relevant and irrelevant labels into a space where they are separated by a margin. $d_+$ is the average distance between the transformed positive word vectors and the origin. See texts for details.}
\label{fig:loss}
\end{figure}

The objective of the deep transformation is to separate the relevant and irrelevant labels of the given image. Specifically, we wish to derive a transformation matrix $A$ by giving an image $\mathcal{I}$ to the CNN model $f_\theta$:
\begin{equation}
	f_\theta(\mathcal{I}) = A.
\end{equation}
The matrix maps labels from a $d$-dim word vector space into a $k$-dim Euclidean space, $w \in \mathbb{R}^d \to w' \in \mathbb{R}^k$, where the relevant labels aggregate around a canonical point (i.e. the origin) and the irrelevant labels scatter far from it.

In multilabel image classification we have a training dataset of pairs $(\mathcal{I}, \{p_i\})$, where each training image $\mathcal{I}$ has several positive labels $\{p_i\}$. We randomly choose other labels $\{n_j\}$ (40 in the experiments) as negatives. The labels are represented by the $d$-dim word vectors (detailed in Section \ref{subsec:word}). The goal is to learn a transformation matrix $A$ from $\mathcal{I}$ so that the distance between the transformed positive word vectors and the origin is smaller than that of negative ones:
\begin{equation}
   \| Ap_i \|_2 < \| An_j \|_2.
\end{equation}   
\noindent
Based on this intuition we define a hinge rank loss $L$ (similar to ~\cite{DeViSE_0}) for a training tuple $(\mathcal{I}, \{p_i\}, \{n_j\})$ as  
\begin{equation}
L = \sum_{j} \max(0, m + \frac{1}{|p_i|}\sum_{i} \| Ap_i \|_2 - \| An_j \|_2),	
\label{equ:loss}
\end{equation}
\noindent
where $m$ is a margin that is enforced between transformed positive and negative word vectors. Note the equation \ref{equ:loss} is a sum of individual losses for negative labels $\{n_j\}$. For each negative label, the loss is zero if $\| An_j \|_2$ is greater by a margin than the average norm $\frac{1}{|p_i|}\sum_{i} \| Ap_i \|_2$. Conversely, if the margin between the norm of negative label and the average norm of the positive labels is violated, the loss is proportional to the amount of violation. This is visualized in Figure \ref{fig:loss}. 

Instead of selecting the closest positive, we use the average norm to eliminate the situation where mislabeled and poorly samples would dominate the loss. Note that the above loss is related to the commonly used triplet loss~\cite{NetVLAD_67, OverFeat, NetVLAD_87, NetVLAD_0}, but it is adapted to multilabel image classification using the formulation given in Section \ref{subsec:formulation}. 

It is worth mentioning that in a special case where we have only a single-label image dataset for training and the transformed dimension ($k$) is set to 1, our model maps the 4,096-dim representation at the top of the visual model into the $d$-dim representation of the word model. This is identical to the behavior of the DeViSE model~\cite{DeViSE_0}, except that DeViSE applies the dot-product similarity to produce the ranking.

\subsection{Word embeddings}
\label{subsec:word}

\begin{figure*}[t]
\begin{center}
%\fbox{\rule{0pt}{2in} \rule{0.9\linewidth}{0pt}}
 \includegraphics[width=0.45\linewidth]{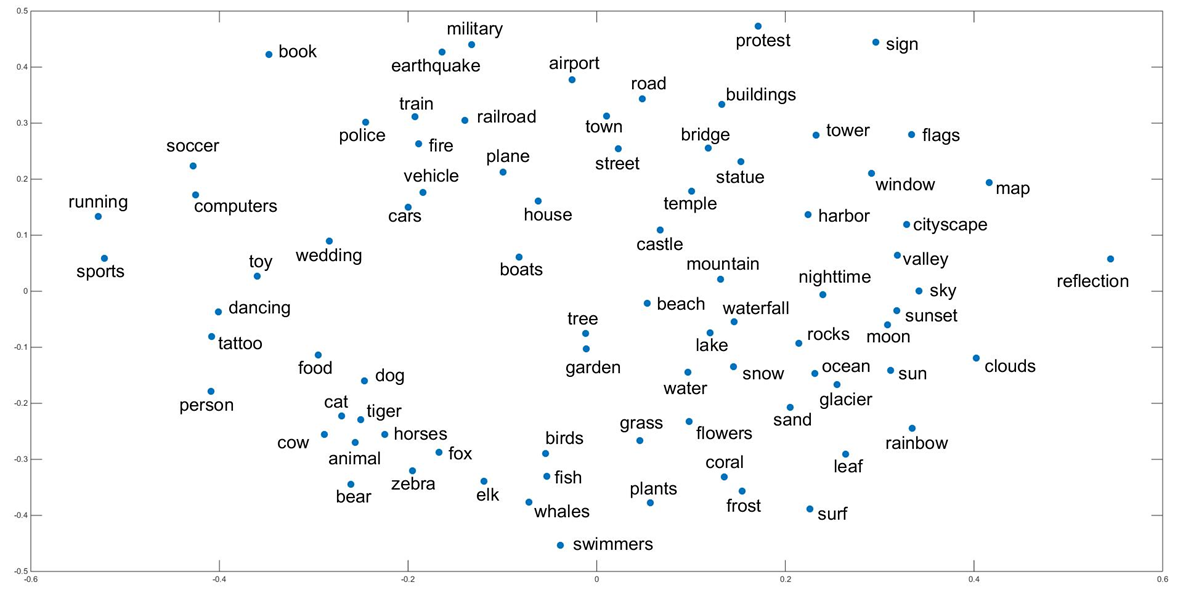}
\end{center}
   \caption{Visualization using t-SNE~\cite{t-SNE} of the label embeddings learned by the word2vec model. These labels are the 81 concepts defined in the NUS-WIDE dataset.}
\label{fig:word}
\end{figure*}

Vector space models (VSMs) represent words in a continuous vector space where semantically similar words are embedded nearby each other. In particular, the skip-gram model introduced by Mikolov \textit{et al.}~\cite{word2vec, DeViSE_14} has been shown to efficiently learn semantically meaningful vector representations of words from a large text corpus. The model learns to represent each word as a fixed length embedding vector by predicting source context words from the target words. Because synonyms tend to appear in similar contexts, the objective function described in~\cite{word2vec} drives the model to learn similar vectors for semantically related words.

In the implementation, we used word2vec~\cite{word2vec} with the skip-gram model of 300-dim embeddings (i.e. $d$ = 300) and trained the word embedding model on the Google News dataset (containing about 100 billion words). Figure \ref{fig:word} displays the visualization of the learned vectors of the 81 concepts defined in the NUS-WIDE dataset~\cite{NUS-WIDE}. The vectors capture some general, and in fact quite useful, semantic information about words and their relationships to one another. For example, the labels of animals (e.g., \textit{dog}, \textit{cat}, \textit{tiger}, \textit{cow}, \textit{horses}, \textit{bear}, \textit{zebra}, \textit{fox}, \textit{elk} and etc.) are gathered around the third quadrant of the figure.

\subsection{Inference}
\label{subsec:inference}

The label prediction of a test image using the proposed image transformation model is trivial. Let $W = \{w_i\}$ denotes the vector representations of the label set. The CNN model takes an input image $\mathcal{I}$ and returns a transformation matrix $A$. For each $w_i$, we calculate $w'_i = A w_i$ and rank the labels according to the L2-norm $\| w'_i \|_2$, i.e. the distance between $w'_i$ and the origin in the new space. Labels with a small distance are retrieved. 

One nice thing about the model is that the label set $W$ does not necessarily contain the labels used in training. The model has the potential to perform zero-shot classification over the unseen labels, because of utilizing word embeddings where unseen and seen labels are in the same vector space. However, zero-shot learning is not the focus of this study. We leave this point for further investigation. 

\subsection{Training details}
\label{subsec:training}

The CNN model was pre-trained on a large-scale single-label image dataset---ImageNet~\cite{ImageNet}. We further trained the network on the target multilabel dataset (e.g., NUS-WIDE~\cite{NUS-WIDE}) with the loss function described in Section \ref{subsec:loss}.

We used Adaptive Moment Estimation (Adam) with momentum 0.9 for 12,000 iterations. We augmented the data by mirroring. The learning rate was set to $10^{-6}$ and was gradually decreased. Training time for a single epoch was around 3 seconds, and training the model roughly took 10 hours. The runtime was reported running on a machine with an Intel Core i7-7700 3.6-GHz CPU, NVIDIA's GeForce GTX 1080 and 32 GB of RAM. The transformed dimension $k$ was set to 100 in the experiments.

\section{Experiments}
\label{sec:exp}

This section presents the experimental results. We compare our approach with several state-of-the-art methods on the large-scale NUS-WIDE dataset~\cite{NUS-WIDE}. We also examine how the transformed dimension $k$ affects the classification performance and interpret the behavior of the model.

\subsection{Experimental settings}
\label{subsec:setup}

\textbf{Dataset.} We evaluated the proposed method on the NUS-WIDE dataset~\cite{NUS-WIDE}. It contains 269,648 images collected from Flickr in the original release. We were able to retrieve only 171,144 images of this dataset because some images were either corrupted or removed from Flickr. We followed the separation guideline from NUS-WIDE and split the dataset into a training set with 102,405 images and a test set with 68,739 images. In each set, the average number of labels per image is 2.43.

NUS-WIDE releases three sets of tags associated with the images. The most widely used set contains 81 concepts, which were carefully chosen to be representative of the Flickr tags and were manually annotated. Therefore, the 81-concepts annotations are much less noisy than those directly corrected from the web. This 81-concepts set is usually used as the ground-truth for benchmarking different multilabel image classification methods. %The second and the third sets of labels are both harvested from Flickr. There are 1,000 popular Flickr tags in the second set and 5,018 raw tags in the third. 

\textbf{Evaluation protocols.} We employed the precision and recall as the evaluation metrics. For each image, the top-$k$ ranked labels are compared to the ground truth labels. The precision is the number of correct labels divided by the number of machine-generated labels. The recall is the number of correct labels divided by the number of ground truth labels.

Following previous researches ~\cite{WARP_0, MIE_0, CNN-RNN_0}, we computed the per-class precision (C-P), overall precision (O-P), per-class recall (C-R) and overall recall (O-R). The average is taken over all classes for computing C-P and C-R, and is taken over all testing examples for computing O-P and O-R. We also reported the F1 score, which is the geometrical average of the precision and the recall.

\subsection{Results}
\label{subsec:results}

\begin{table*}
\begin{center}
\begin{tabular}{|l|c|c|c|c|}
\hline
& word model & RNN & region proposals & CNN architecture \\
\hline\hline
WARP~\cite{WARP_0}       &            &            & \checkmark & AlexNet \\
CNN-RNN~\cite{CNN-RNN_0} & \checkmark & \checkmark &            & VGG-16 \\ 
MIE~\cite{MIE_0}         & \checkmark &            &            & GoogleNet \\
Ours                     & \checkmark &            &            & AlexNet \\
\hline
\end{tabular}
\end{center}
\caption{Summary of the methods under comparison}
\label{tab:methods}
\end{table*}

\begin{table*}
\begin{center}
\begin{tabular}{|l|c|c|c|c|c|c|}
\hline
Method & C-P & C-R & C-F1 & O-P & O-R & O-F1  \\
\hline\hline
WARP~\cite{WARP_0}    & 31.7\%	& 35.6\%	& 33.5\% & 48.6\%	& 60.5\%	& 53.9\%	\\
CNN-RNN~\cite{CNN-RNN_0} & 40.5\%	& 30.4\%	& 34.7\% & 49.9\%	& 61.7\%	& 55.2\%	\\ 
MIE~\cite{MIE_0}     & 37.7\%	& 40.2\%	& 38.9\% & 52.2\%	& 65.0\%	& 57.9\%  \\
Ours            & 36.7\%	& 41.2\%	& 38.9\% & 51.8\%	& 63.8\%	& 57.2\%  \\
\hline
\end{tabular}
\end{center}
\caption{Multilabel image classification results on NUS-WIDE with 3 predicted labels per image. The number of training and testing images used in our method are 102,405 and 68,739 and those in other methods are 150,000 and 59,347.}
\label{tab:results_k3}
\end{table*}

\begin{table*}
\begin{center}
\begin{tabular}{|l|c|c|c|c|c|c|}
\hline
Method & C-P & C-R & C-F1 & O-P & O-R & O-F1 \\
\hline\hline
WARP~\cite{WARP_0}    & 	22.3\%	& 52.0\%	& 31.2\% & 36.2\%	& 75.0\%	& 48.8\%	\\
MIE~\cite{MIE_0}     & 	28.3\%	& 59.8\%	& 38.4\% & 39.0\%	& 80.9\%	& 52.6\%	\\
Ours            & 	27.5\%  & 58.5\%  & 37.4\% & 38.8\% & 79.7\%  & 52.2\%  \\
\hline
\end{tabular}
\end{center}
\caption{Multilabel image classification results on NUS-WIDE with 5 predicted labels per image. The number of training and testing images used in our method are 102,405 and 68,739 and those in other methods are 150,000 and 59,347.}
\label{tab:results_k5}
\end{table*}

We compared the proposed method with recent CNN-based competitive methods.

\begin{itemize}
   \item WARP~\cite{WARP_0}: WARP uses the AlexNet trained with weighted approximate ranking (the WARP loss)~\cite{Wsabie}. It specifically optimizes the top-$k$ accuracy for classification by using a stochastic sampling approach. 
 \item CNN-RNN~\cite{CNN-RNN_0}: This framework incorporates Long Short-Term Memory Networks (LSTM)~\cite{LSTM} with the 16 layers VGG network~\cite{VGGNet} to model label dependency. 
 \item MIE~\cite{MIE_0}: MIE applies the Fast R-CNN~\cite{Fast-R-CNN} to construct region proposals and uses a fully connected layer to embed each image subregion into the semantic space.  
\end{itemize}

Table \ref{tab:methods} summarizes the methods. Note that our training set contains 102,405 images, which occupies only 68.27\% of the set used in these competing methods (150,000 images).  

We reported the experimental results with 3 and 5 predicted labels for each image in Table \ref{tab:results_k3} and Table \ref{tab:results_k5}, respectively. The proposed method consistently outperformed WARP in terms of all measurements. Since both methods enforced positive labels to be top ranked, the performance gain (C-F1: 6.2\%, O-F1: 3.4\% in top-5 prediction) may be obtained by using a word model that provided a priori knowledge about the labels. 

In comparison with RNN-CNN, both methods modeled image-label and label-label dependencies. The proposed method performed slightly better than RNN-CNN (C-F1: 4.2\%, O-F1: 2\% in top-3 prediction) despite a much simpler network architecture was used. Finally, the proposed method had a comparable performance with MIE. However, the design principles of these methods were completely different. MIE used GoogleNet which was deeper and more complex than the AlexNet used in our model. MIE required additional computations to extract semantically meaningful sub-regions from one image, while the proposed method took a global approach. MIE modeled the region-to-label correspondence and ours modeled that between an image and its label set. The proposed method was much simpler than MIE.

\subsection{Empirical analysis on the transformed dimension}
\label{subsec:k_value}

\begin{figure}
    \centering
    \begin{subfigure}[b]{0.35\textwidth}
        \includegraphics[width=\textwidth]{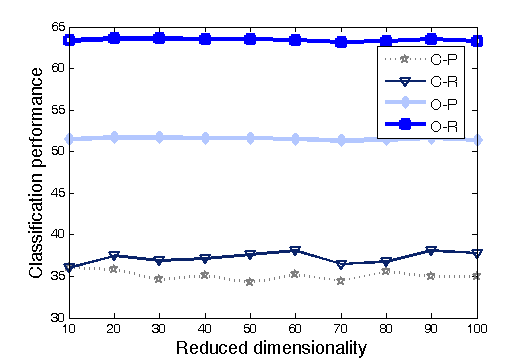}
        \caption{top-3 prediction}
        \label{fig:top3_k}
    \end{subfigure}
    ~ %add desired spacing between images, e. g. ~, \quad, \qquad, \hfill etc. 
      %(or a blank line to force the subfigure onto a new line)
    \begin{subfigure}[b]{0.35\textwidth}
        \includegraphics[width=\textwidth]{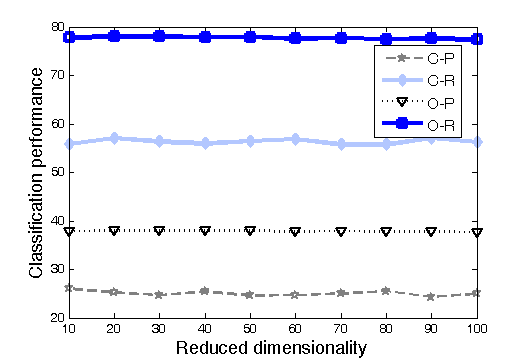}
        \caption{top-5 prediction}
        \label{fig:top5_k}
    \end{subfigure}
    \caption{Effect of the transformed dimension to the classification performance.}\label{fig:k_value}
\end{figure}

Recall that we learn a transformation matrix from an image that maps the labels from the word space to a $k$-dim Euclidean space. Now we examine the design choice in determining the dimension ($k$). In this experiment we trained 10 models by uniformly sampling $k$ from 10 to 100. Each model was trained with 1000 iterations. 

Figure \ref{fig:k_value} shows the classification performance of these models in top-3 and top-5 predictions. The classification performance is rather stable no matter which $k$ value is used. The determination of the $k$ value has little effect on the overall classification performance.

\subsection{Model interpretation}
\label{subsec:interpretation}

\begin{table}
\begin{center}
\begin{tabular}{|l|c|c|c|c|}
\hline
Method & C-P & C-R & O-P & O-R  \\
\hline\hline
%WARP~\cite{WARP_0}    & 31.7\%	& 35.6\%	& 48.6\%	& 60.5\%	\\
%CNN-RNN~\cite{CNN-RNN_0} & 40.5\%	& 30.4\%	& 49.9\%	& 61.7\%	\\ 
%MIE~\cite{MIE_0}     & 37.7\%	& 40.2\%	& 52.2\%	& 65.0\%  \\
%\hline
Voting (top 1)  & 23.2\%	& 31.9\%	& 47.9\%	& 59.0\%  \\
Voting (top 3)  & 29.1\%	& 36.3\%	& 50.6\%	& 62.2\%  \\
Voting (top 5)  & 31.8\%	& 37.4\%	& 51.1\%	& 62.9\%  \\
\hline
Full            & 36.7\%	& 41.2\%	& 51.8\%	& 63.8\%  \\
\hline
\end{tabular}
\end{center}
\caption{Top 3 prediction results. The model can be viewed as a combination of $k$ classifiers with shared features.}
\label{tab:interpretation}
\end{table}

The proposed model can be viewed as a combination of $k$ classifiers with shared CNN features to produce a powerful ``committee.'' Recall that we obtain a $k \times d$ transformation matrix $A$ from an image via CNN. Each $d$-dim row vector in this matrix can be interpreted as a principal direction in the original word vector space, along which the labels are ranked. Simultaneously training $k$ CNNs is costly and may cause overfitting, and we avoid these problems using a shared CNN---all classifiers share the same image features. We use a single CNN to implement an assembly of $k$ classifiers.

This strategy leads to $k$ powerful and complementary classifiers. To illustrate this point, we individually inspected the outputs of each classifier and retained only top $N$ labels of a classifier. Next, we used a simple voting scheme to aggregate the results of all classifiers. For example, we obtained $k$ labels (may be repetitive) when $N$ was set to 1, from which we retrieved frequent labels as the final output.

Table \ref{tab:interpretation} shows the classification performances in top-3 prediction with various $N$ values. By using a small $N$ (i.e. $N$ = 3) the combination of the $k$ classifiers outperformed WARP~\cite{WARP_0} and CNN-RNN~\cite{CNN-RNN_0}. The classification performance was further boosted when all results are used jointly. 

\begin{figure}[t]
\begin{center}
%\fbox{\rule{0pt}{2in} \rule{0.9\linewidth}{0pt}}
 \includegraphics[width=0.85\linewidth]{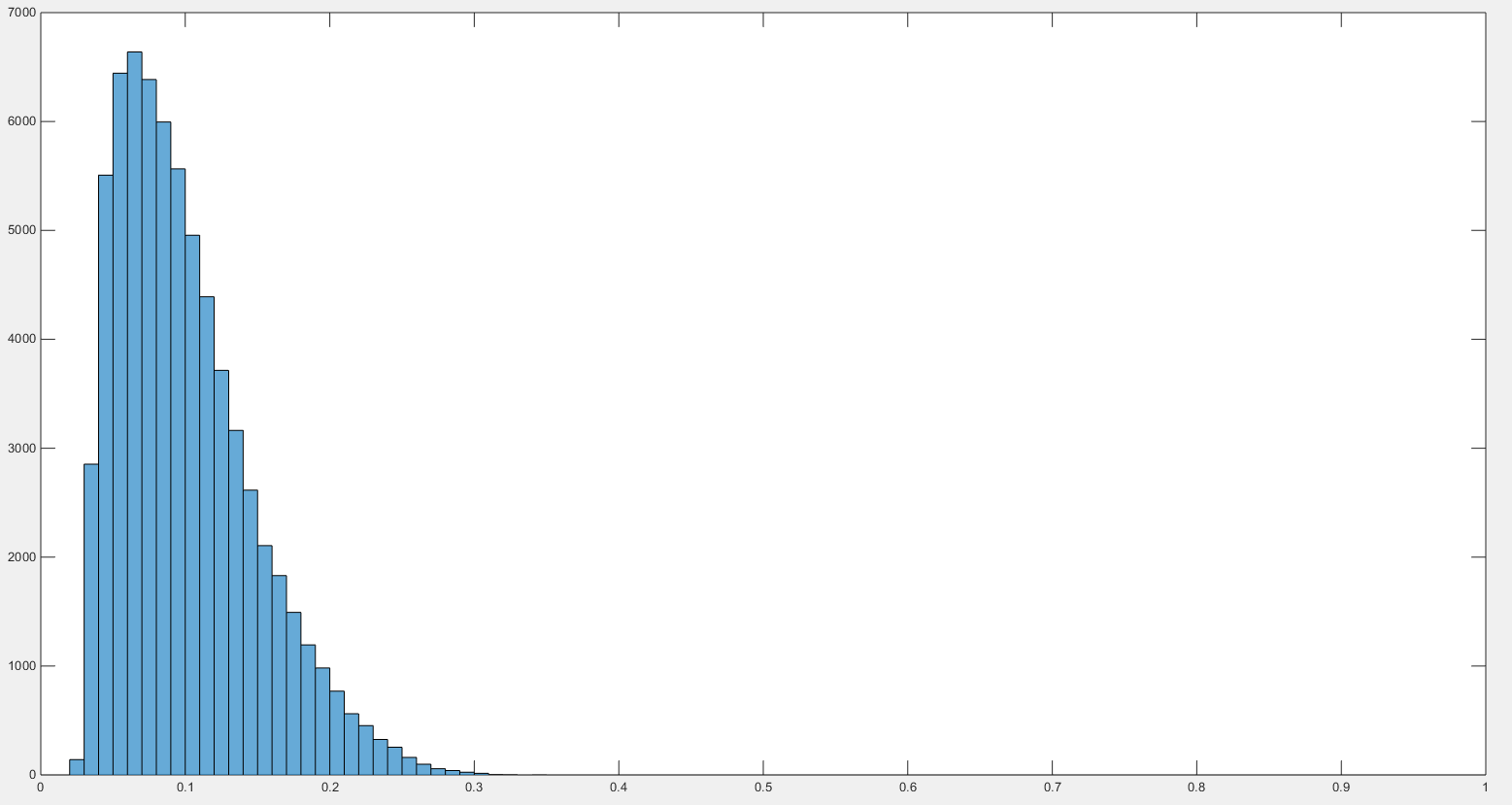}
\end{center}
   \caption{Distribution of the average Jaccard coefficients. The output labels of the $k$ classifiers are different.}
\label{fig:similarity}
\end{figure}

Next, we investigated the similarities among the outputs of the individual classifiers. In this empirical analysis, we obtained the top 5 predicted labels from each classifier. The Jaccard coefficient---defined as the size of the intersection divided by the size of the union of two sets---was used to compute the pair-wise similarity of two label sets. For each test image, we reported the average Jaccord coefficient of the $\binom{k}{2}$ pairs. Figure \ref{fig:similarity} shows the histogram of the average Jaccord coefficients computed using the test set. The mean value is 0.1001 and the standard deviation is 0.1288, indicating that the outputs of the classifiers are very different. As shown in Table \ref{tab:results_k3}, combining these classifiers led to a powerful committee. 

This interpretation relates the proposed method to Fast0Tag~\cite{Fast0Tag}, which aims to learn a mapping function between the visual space and the word vector space. This approach can be viewed as a special case of our method by setting $k$ to 1. 

%\begin{figure}[t]
%\begin{center}
% \includegraphics[width=0.95\linewidth]{./figures/visualization.png}
%\end{center}
%   \caption{}
%\label{fig:visualization}
%\end{figure}

\section{Conclusion and future work}
\label{sec:conclusion} 

We have extended single-label visual-semantic embedding models for multilabel image classification. The complex image-to-label and label-to-label dependencies are modeled via a simple infrastructure involving only a single CNN as the visual model. In particular, a new learning paradigm is developed: we learn a transformation---rather than a representation---from an image, with the objective of optimizing the separation of the image's relevant and irrelevant labels. Fast and accurate prediction of labels can be achieved by simply performing a linear transformation on the word vectors.

One future research direction we are pursuing is to extend the method for zero-shot prediction, in which test images are assigned with unseen labels from an open vocabulary. This would take full advantage of the word model---unseen labels are in the same vector space as the seen labels for training. Another direction is to explore the learning of nonlinear transformation, which may better exploit higher order dependencies among labels.

{\small
\bibliographystyle{ieee}
\bibliography{egbib}
}

\end{document}